\def\year{2022}\relax
\documentclass[letterpaper]{article} 
\usepackage{styles/aaai22}  
\usepackage{times}  
\usepackage{helvet}  
\usepackage{courier}  
\usepackage[hyphens]{url}  
\usepackage{graphicx} 
\def\UrlFont{\rm}  
\usepackage{natbib}  
\usepackage{caption} 
\DeclareCaptionStyle{ruled}{labelfont=normalfont,labelsep=colon,strut=off} 
\usepackage{algorithm}
\usepackage{algorithmic}
\usepackage{amsmath}
\usepackage{amsfonts}
\usepackage{amsthm}
\usepackage{xcolor}
\usepackage{newfloat}
\usepackage{listings}
\floatstyle{ruled}
\newfloat{listing}{tb}{lst}{}

\floatname{listing}{Listing}

\DeclareMathOperator{\EX}{\mathbb{E}}

\title{On the Use and Misuse of Absorbing States in Multi-agent Reinforcement Learning}

\newcommand{\eg}{\emph{e.g.,} }
\newcommand{\ie}{\emph{i.e.,} }
\author{Andrew Cohen\equalcontrib\footnote{Corresponding author andrew.cohen@unity3d.com},
        Ervin Teng\equalcontrib,
        Vincent-Pierre Berges\equalcontrib,
        Ruo-Ping Dong,
        Hunter Henry,\\
        Marwan Mattar,
        Alexander Zook,
        Sujoy Ganguly
  }
\affiliations{
    Unity Technologies
}

\begin{document}
\maketitle

\begin{abstract}
The creation and destruction of agents in cooperative multi-agent reinforcement learning (MARL) is a critically under-explored area of research. Current MARL algorithms often assume that the number of agents within a group remains fixed throughout an experiment. However, in many practical problems, an agent may terminate before their teammates. This early termination issue presents a challenge: the terminated agent must learn from the group's success or failure which occurs beyond its own existence. 
We refer to propagating value from rewards earned by remaining teammates to terminated agents as the \emph{Posthumous Credit Assignment} problem. 
Current MARL methods handle this problem by placing these agents in an \emph{absorbing state} until the entire group of agents reaches a termination condition. Although absorbing states enable existing algorithms and APIs to handle terminated agents without modification, practical training efficiency and resource use problems exist. 

In this work, we first demonstrate that sample complexity increases  with the quantity of absorbing states in a toy supervised learning task for a fully connected network, while attention is more robust to variable size input. Then, we present a novel architecture for an existing state-of-the-art MARL algorithm which uses attention instead of a fully connected layer with absorbing states. Finally, we demonstrate that this novel architecture significantly outperforms the standard architecture on tasks in which agents are created or destroyed within episodes as well as standard multi-agent coordination tasks.


\end{abstract}

\section{Introduction}
In many real-world scenarios, agents must cooperate to achieve a shared objective. In these settings, single-agent reinforcement learning (RL) methods can fail or perform sub-optimally for various reasons, such as the partial observability inherent in multi-agent systems, exacerbated by increasing numbers of agents. Multi-agent reinforcement learning (MARL) promises to address these issues using the paradigm of \emph{decentralized execution} and \emph{centralized training}~\cite{Lowe2017}. In this paradigm, agents act using \emph{local} observations, but all \emph{globally} available information is used during training.

The MARL literature~\cite{Lowe2017, Foerster18,Long20, Iqbal21} often assumes that we will train a fixed number of agents. However, this is unsuitable for many practical applications of MARL. For instance, agents in a team-based video game may ``spawn" (\ie be created) or ``die" (\ie terminate before the other agents) within a single episode. Similarly, robots operating as a team may run out of battery, requiring that they terminate their trajectories before their teammates. In general, an agent can terminate early, meaning it no longer influences the environment or other agents mid-episode. Furthermore, one may also acquire additional agents mid-episode.

Typically, existing algorithms handle these situations by placing inactive agents in \emph{absorbing states}. An agent remains in an absorbing state, irrespective of action choice, until the entire group of agents reaches a termination condition. Absorbing states enable existing algorithms to train cooperative agents to solve tasks with early termination without any architectural changes and also simplify environment and multi-agent API implementations. Furthermore, absorbing states enable decentralized POMDPs~\cite{Oliehoek} and Markov Games~\cite{Littman94} to represent tasks with early termination without modification.   

However, absorbing states introduce practical problems in training efficiency and resource use. Specifically, when using neural networks as function approximators, absorbing states introduce elements into the input distribution that, by the necessities of their construction, make the target function more challenging to approximate. Furthermore, absorbing states are not a scalable solution for large numbers of agents. Depending on the problem, a non-insignificant amount of resources may be used for agents that do not influence the environment. In an extreme case, if a group has more than $50\%$ attrition, more resources are consumed for communication and storage for non-influential agents. These underutilized resources are of particular concern when there are strict resource constraints.

The critical challenge posed by the early termination of an agent is credit assignment--which we call Posthumous Credit Assignment. Agents removed from the environment will not experience any rewards given to the group after termination. As such, they will not learn if their actions before termination were valuable to the group. Absorbing states solve this by creating a pathway through the state space by which to propagate value from beyond an agent's early termination. 


In this work, we present a novel architecture which uses attention instead of a fully connected layer with absorbing states for the state-of-the-art MARL algorithm COunterfactual Multi-Agent Policy Gradients (COMA)~\cite{Foerster18}. We refer to our proposed architecture as \textbf{M}ulti-{\bf A}gent \textbf{PO}sthumous \textbf{C}redit \textbf{A}ssignment (MA-POCA).
MA-POCA naturally handles agents that are created or destroyed within an episode but share a reward function. Working within the centralized training, decentralized execution framework, we need only enable the \emph{critic} to handle a changing number of agents per timestep. By applying a self-attention mechanism~\cite{Vaswani2017} to only the active agent information before the critic, MA-POCA can scale to an arbitrary number of agents. 
Furthermore, the attention mechanism allows the critic to attribute the future expected value of the group to states with terminated agents \emph{without absorbing states}. Lastly, the attention mechanism enables the implementation of the counterfactual baseline~\cite{Foerster18} for agents with both continuous and discrete action spaces. 


This work has three main contributions.
\begin{itemize}
    \item We demonstrate that sample complexity increases  with the quantity of absorbing states on a toy supervised learning task for a fully connected network, while attention is more robust to variable size input.
    \item We present a novel architecture, MA-POCA, which propagates rewards earned by remaining teammates to terminated agents without the use of absorbing states. Furthermore, because it does not rely on a fixed number of agents, MA-POCA also supports the \emph{creation} of new agents during an episode.
    \item We present experiments on two standard multi-agent coordination tasks and two novel tasks in which agents can spawn or die. We show that MA-POCA provides improvement on the former and significantly outperforms the baselines on the latter.
\end{itemize}


\section{Preliminaries}\label{prelim}

\subsection{MARL notation}
The setting we consider is a decentralized-POMDP~\cite{Oliehoek} defined by: $(N, \mathcal{S}, \mathcal{O}, \mathcal{A}, P, r, \gamma )$ where $N \ge 1$ is the number of agents and $\mathcal{S}$ is the state space of the environment. 
$\mathcal{O}$ is the joint observation space of all agents $\mathcal{O}:= \mathcal{O}^1 \times ... \times \mathcal{O}^N$ where $\mathcal{O}^i$ is the observation space of agent $i$. 
At time $t$, the environment is in state $s_t \in \mathcal{S}$ and $o^i_t \in \mathcal{O}^i$ is the local observation of agent $i$ which is correlated with $s_t$. 
The environment state may contain information that is not available locally to any agent such as the total number of agents that are currently acting. 
$\mathcal{A}$ is the joint action space of all agents $\mathcal{A}:= \mathcal{A}^1 \times ... \times \mathcal{A}^N$ where $\mathcal{A}^i$ is the action space of agent $i$. 
Note, the observation and action spaces for different agents do not need to be equal. 
Additionally, we use bold vectors to represent the joint quantities over agents \eg a joint action $\boldsymbol{a} = (a^1,\ ...\ ,\ a^N)$ or joint observation $\boldsymbol{o} = (o^1,\ ...\ ,\ o^N),$ where $\boldsymbol{a} \in \mathcal{A}$ and $\boldsymbol{o} \in \mathcal{O}$.

$P: \mathcal{S} \times \mathcal{A} \times \mathcal{S} \rightarrow [0,1]$ is the transition function where $P(s'|s, \boldsymbol{a})$ is the probability that the environment transitions to state $s'$ given the current state $s$ and joint action $\boldsymbol{a} \in \mathcal{A}$. $r : \mathcal{S} \times \mathcal{A} \rightarrow \mathbb{R}$ is the shared reward function where $r(s, \boldsymbol{a})$ is the reward received by all agents when the joint action $\boldsymbol{a} \in \mathcal{A}$ is taken and the environment is in state $s \in \mathcal{S}$. 

\subsection{Centralized Training, Decentralized Execution}
In this work, we consider the Independent Actor with Centralized Critic (IACC) learning framework~\cite{Lyu21} wherein a critic trained on joint information is used to update a set of independent actors in an actor-critic architecture~\cite{Konda00}. Independent Actor-Critic (IAC), which trains an independent critic and policy for each agent using only local information, and the Joint Actor-Critic (JAC), which trains a single joint policy and a joint critic, are competing approaches. In general, IAC does not perform well in tasks that require significant coordination because of the partial observability in using only local observations. Additionally, JAC is not practical in real world scenarios as a joint policy needs access to all agent observations at once to generate actions, essentially presuming perfect communications between agents and the policy node.

Let $\pi_i,\ 1\le i \le N$ represent the policy of each independent actor. Given the environment state $s_t$ and corresponding joint observation $\boldsymbol{o_t}$ and action $\boldsymbol{a_t}$, the joint policy $\boldsymbol{\pi}(\boldsymbol{a_t} | \boldsymbol{o_t})$ can be factored as $\boldsymbol{\pi}(\boldsymbol{a_t} | \boldsymbol{o_t}) = \prod_i \pi_i(a^i_t | o^i_t)$ since the agents act independently on local observations.

The centralized state value function for state $s_t$ is defined as
\begin{align}\label{csvf}
    V^{\boldsymbol{\pi}}(s_t) = \mathbb{E}_{\boldsymbol{\pi}}\left[\sum_{l=0}^\infty \gamma^l r(s_{t+l}, \boldsymbol{a_{t+l}})\right]
\end{align}
and the centralized state-action value function as 
\begin{align}\label{csavf}
    Q^{\boldsymbol{\pi}}(s_t,\boldsymbol{a_t}) = r(s_{t}, \boldsymbol{a_{t}}) +  \mathbb{E}_{\boldsymbol{\pi}}\left[\sum_{l=1}^\infty \gamma^l r(s_{t+l}, \boldsymbol{a_{t+l}})\right]
\end{align}

\subsection{Counterfactual Baselines}
\emph{Counterfactual baselines} leverage difference rewards~\cite{Wolpert02} and introduce a per-agent baseline such that the advantage reflects the individual agent's contribution to the total reward~\cite{Foerster18}. Formally, the state action value function with the action of the individual agent marginalized out is used to compute the baseline
\begin{align}\label{counterfac}
     b_i(s,\boldsymbol{a}) = \mathbb{E}_{a' \sim \pi_i(\cdot|o_i)}[Q^{\boldsymbol{\pi}}(s,(\boldsymbol{a^{-i}},a'))]
\end{align}
where $\boldsymbol{a^{-i}}$ is the joint action without the $i$'th entry.
Then, the advantage of agent $i$ is
\begin{align}
    \text{Adv}_i = Q^{\boldsymbol{\pi}}(s,\boldsymbol{a}) - b_i(s,\boldsymbol{a})
\end{align}
and the update for agent $i$ is
\begin{multline}\label{policy_update}
     \nabla_{\theta_i} J(\theta_i) =\\ \EX_{\substack{s\sim\rho^\pi\\ a^i\sim\pi_i}}[\nabla_{\theta_i} \log\pi_i(a^i| o^i)(Q^{\boldsymbol{\pi}}(s,\boldsymbol{a}) - b_i(s,\boldsymbol{a}))]
\end{multline}
Using this as the advantage function provides a shaped reward per agent that addresses the challenge of determining of how much an individual agent contributed to the shared reward of the group. Additionally, with the use of the counterfactual baseline, gradient descent still converges to the locally optimal policy~\cite{Foerster18}.

\section{Challenges of Early Terminating Agents}
In this section, we introduce the Posthumous Credit Assignment problem and discuss how it fits into decentralized POMDP framework. To the author's knowledge, this is first time the posthumous credit assignment problem has been explicitly mentioned in the literature. Then, we discuss how the decentralized POMDP framework can be extended to destroying agents via the use of an \emph{absorbing state}. Absorbing states appear in the MARL literature~\cite{samvelyan19smac,Yu2021}, but we provide an explicit discussion of them which the literature lacks. Finally, we discuss practical and theoretical issues introduced by absorbing states. 

\subsection{Posthumous Credit Assignment}
In cooperative settings with shared rewards, an agent acts to maximize the expected future reward of the group. There are scenarios in which an individual agent's current actions enable the group to obtain reward at a later timestep, but result in the immediate termination of the agent itself (\eg a self-sacrificial event). From the perspective of a reinforcement learning agent, it has been removed from the environment and therefore will no longer receive the reward its group may obtain later. Additionally, the agent is unable to observe the state of the environment at the time the group receives the reward. Therefore, an agent must learn to maximize rewards that it cannot experience, presenting a critical credit assignment problem. We call this the \emph{Posthumous Credit Assignment} problem.

\subsection{Absorbing States}\label{sec:abs_states}
The decentralized POMDP framework is equipped to model tasks with the posthumous credit assignment problem via absorbing states. For each agent, let $o_i^{abs} \in \mathcal{O}_i$ be a unique absorbing state which agent $i$ will occupy after it has reached a termination state and is no longer active in the environment. Note, this absorbing state needs to be \emph{per agent} as agents with different observation spaces will enter absorbing states of different dimensions. Once agent $i$ has entered $o_i^{abs}$, it will remain there, irrespective of actions, until the group has reached a termination condition and all agents reset to a new initial state. Thus, the following is true for $o_i^{abs}$
\begin{align*}
    p(o_i^{abs}| o_i^{abs}, a_i) = 1,\ \forall a_i \in \mathcal{A}_i. 
\end{align*}

Additionally, when agent $i$ is in state $o_i^{abs}$, the transition function will be independent of that agent's actions. Formally,
\begin{align*}
    P(s'|s, \boldsymbol{a}) = P(s'|s, \boldsymbol{a^{-i}})
\end{align*}
where $\boldsymbol{a^{-i}}$ is the joint action without the $i$'th entry. Thus, the introduction of an absorbing state transforms the original problem into a standard decentralized POMDP without the issue of posthumous credit assignment.

However, though representing the setting of dying and spawning agents with the decentralized POMDP formalism and absorbing states is straightforward, issues arise when using absorbing states in practice. We argue that it is best to avoid absorbing states in general. The two major concerns are, 
\begin{itemize}
    \item the absorbing state representation complicates the learning dynamics of neural network-based function approximators;
    \item the complexity introduced and wasted computational resources consumed for agents that no longer impact the environment.
\end{itemize} 

\subsubsection{Absorbing State Representation}
The absorbing state representation is non-trivial since, in most algorithms, a function approximator, like a neural network, will ingest these state representations. Thus, the actual values and structure of the state are important as they act as input features. Additionally, the absorbing state must be disjoint from the observation space accessible by active agents. Otherwise, we would introduce (additional) partial observability since the same values could represent an active or inactive agent.

\begin{figure}
    \centering
    \includegraphics[width=.48\textwidth]{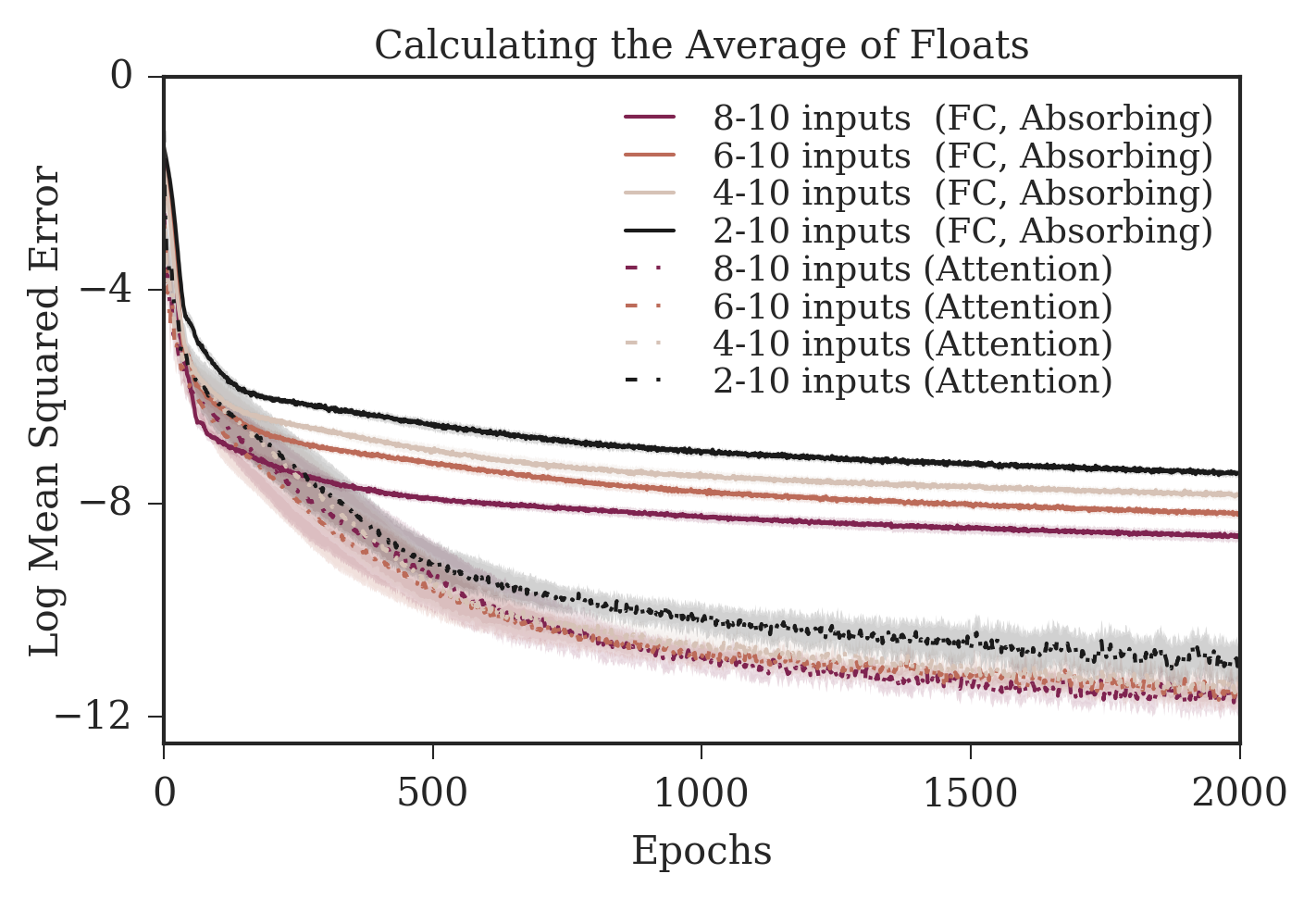}
    \caption{The sample efficiency of learning to compute the mean of a varying number of floats depends on the quantity of absorbing states as well as the representation of the input to the network. Attention outperforms absorbing states in both performance and robustness to larger variation in the input. Curves are mean and $95\%$ confidence interval over $20$ seeds. }
    \label{fig:abs_ablate}
\end{figure}

As a case study, we present a toy numerical example in Figure~\ref{fig:abs_ablate} to illustrate that fully connected layers with absorbing states are not a sample efficient input representation compared to attention networks. We train a neural network to estimate the mean of a variable number of uniformly sampled floats (up to $10$) in the range $[0.25,0.75]$. We compare two configurations (all hyperparameters are contained in Appendix~\ref{sec:hyperparameters}): 
\begin{itemize}
\item the remaining values are substituted with a fixed absorbing state $o_{abs}$ and the sample is shuffled to simulate agents terminating early in the RL scenario. We train a fully connected network with 2 hidden layers of 32 units with ReLU activation functions; 
\item a self-attention~\cite{Vaswani2017} layer processes the variable input. We use a single layer entity embedding of size 32, a residual self-attention ($RSA$) block followed by a linear transformation into the output space. The implementation details of the RSA block are contained in Appendix~\ref{atten_app}.
\end{itemize}

In Figure~\ref{fig:abs_ablate}, we provide loss curves for learning to compute the mean of 2 through 10, 4 through 10, 6 through 10, and 8 through 10 floats using both fully connected layers with absorbing states and attention. We draw the number of inputs from a uniform random distribution within the ranges for each data point.
In this experiment, the value of the absorbing state is $o_{abs}=0.0$ but the trends are similar for other choices. Note that we \emph{cannot} choose $o_{abs} \in [0.25, 0.75]$ as it will not be possible for the network to learn when a value should or should not be included in the mean. In Appendix~\ref{abs_curves}, we provide additional figures for the different values of $o_{abs} = -1.0,\ 1.0,\ 0.4$ showing that $-1.0$ and $1.0$ are reasonable choices but $0.4$ is not as it is contained in $[0.25, 0.75]$. Additionally, we provide an experiment when the number of absorbing states is fixed per sample to demonstrate that varying the number of floats is more challenging. 

When using absorbing states, we observe that the sample complexity \emph{increases} with the number of absorbing states. The runs with a greater maximum number of absorbing states take longer to converge. We also observe that attention significantly outperforms absorbing states in this task. 




When extrapolating this result to the RL setting, the mappings learned by centralized value functions are much more complicated than this simple numerical example, likely amplifying the issues discussed. Furthermore, in this example, the presence of an absorbing state has exactly one meaning \ie do not include this input in the computation of the mean. However, in a MARL setting, the group's outcome can be positive, negative, or neutral following the early termination of an agent. Then, the variation in outcomes corresponds to high variance return targets for the single absorbing state $o_{abs}$. A possible alternative is to use multiple absorbing states, one for each outcome. However, it may not be possible to know which outcome will follow a given early termination or outcomes may be on a spectrum. 

\subsubsection{Implementation Complexity and Resource Constraints}

From a practical standpoint, absorbing states require both additional implementation complexity and increased resource overhead. These issues arise mainly in two areas: the sizing of the function approximator that represents the centralized value function and the communication and storage of redundant absorbing states.

In order to use absorbing states, we must size the centralized value function approximator to take as input all of the observations from \emph{the absolute maximum} number of agents that can be active in the environment, regardless of how many are active at any given time. If the function approximator is a fully connected neural network, it must have inputs for all possible agents. In addition to adding sample complexity to the learning process, these extra parameters present an unnecessary computational overhead during training. Furthermore, in cases where the maximum number of agents is unknown, \eg when agents' actions can spawn additional agents, we must choose some arbitrarily large value for the maximum number of agents, exacerbating the sample complexity and computational overhead issues of the function approximator. 

We must also consider the computational and resource overhead of the absorbing states themselves. Most implementations of absorbing states (\eg SMAC~\cite{samvelyan19smac}) add them as part of the state returned from the environment. This has the advantage of not requiring explicit knowledge of absorbing states by the algorithm implementation. However, as these absorbing states are treated in the same way as any other states, they must also be processed, stored, and communicated in the same way as other states. In distributed RL architectures which rely on distributed inference workers~\cite{Espeholt2018,Horgan2018}, these states would need to be sent from these workers to the optimizer as part of trajectories, where they would introduce a communication overhead. In addition, they will exist in any buffers, queues, and, in the case of off-policy algorithms, replay stores, taking up unnecessary memory. These issues can be partially mitigated by moving the implementation of absorbing states from the environment to the algorithm (\eg padding states right before they are given as input to the centralized value function), at the cost of making the implementation of the algorithm more complex and less general across environments.

\section{Methods}\label{method}
In this section, we propose a novel architecture for COMA~\cite{Foerster18} called MA-POCA. MA-POCA uses self-attention~\cite{Vaswani2017} over \emph{active} agents in the critic network, thereby addressing the issue of posthumous credit assignment without the need for absorbing states. Additionally, self-attention enables a network architecture that can efficiently compute counterfactual baselines for groups of homogeneous and heterogeneous agents. Note that, though the decentralized POMDP framework requires that the maximum number of agents $N$ is known, the algorithm and network architecture of MA-POCA do not. 

\subsection{MA-POCA}
 
MA-POCA learns a centralized value function to estimate the expected discounted return of the group of agents and a centralized agent-centric counterfactual baseline to achieve credit assignment in the manner of COMA. In architectures that use self-attention, it is common to have entity encoders which map entities to an embedding space before passing through the attention layer~\cite{Baker20}. In our setting, we consider \emph{distinct} observation spaces as entities. For example, if two agents $i,\ j$ share the same observation space $\mathcal{O}_i = \mathcal{O}_j$, corresponding observations will be embedded with the \emph{same encoder}. However, if $\mathcal{O}_i \ne \mathcal{O}_j$, we embed them with different encoders. Furthermore, we consider observations and observation-action pairs as separate entities; observation and actions are concatenated and then embedded.  

Formally, let $g_i: \mathcal{O}_i\rightarrow E$ be an encoding network for observations $o_i \in \mathcal{O}_i$ where $E$ is the embedding space. As stated previously, if $\mathcal{O}_i = \mathcal{O}_j$, then $g_i = g_j$.

\subsection{MA-POCA Value Function}
In this section, we discuss how to estimate the expected discounted return given in Eq.~\ref{csvf} for a group of agents wherein some may terminate early. Recall, in our setting, the number of active agents depends on $t$. Thus, let $k_t$ denote the number of active agents at time step $t$ such that $1 \le k_t \le N$ where $N$ is the maximum number of agents that can be alive at any time. 

In the MARL literature, there are generally two ways that centralized state or state-action value functions are conditioned on state: 
\begin{itemize}
    \item there is a separate vector containing global information (\ie the coordinates of all agents) that is unobserved by any individual agent~\cite{Foerster18, Rashid18};
    \item the joint observation of the agents $\bf{o_t} \in \mathcal{O}$ is used as this is a reasonable approximation to the global state~\cite{Long20, Lowe2017}. 
\end{itemize}
It is also possible to use a hybrid. In this work, we only consider the joint observation of \emph{active agents}, though the former would still require absorbing states or values in some capacity and can be treated by means similarly to what follows.

To handle a varying number of agents per timestep, we first encode the observations of all active agents $g_i(o^i_{t})_{1\le i \le k_t}$ and then pass the encodings through an $RSA$. The $RSA$ block we use is architecturally similar to those used in the vanilla Transformer architecture~\cite{Vaswani2017} but without positional encodings~\cite{Baker20}. For more details on the self-attention mechanism, please see Appendix~\ref{atten_app}. Then, the centralized state value function parameterized by $\phi$ has the form 
\begin{align}\label{mapocavf}
    V_{\phi}(RSA(g_i(o^i_{t})_{1\le i \le k_t}))
\end{align}
and is trained with $TD(\lambda)$~\cite{Sutton1988}
\begin{align}\label{value_objf}
    J(\phi) = (V_{\phi}(RSA(g_i(o^i_{t})_{1\le i \le k_t})) - y^{(\lambda)})^2
\end{align}
where
\begin{multline*}
y^{(\lambda)} = (1-\lambda) \sum_{n=1}^{\infty} \lambda ^ {n-1} G_{t}^{(n)} \\
G_{t}^{(n)} = \sum_{l=1}^n\gamma^{l-1}r_{t+l} + \gamma^n V_\phi(RSA(g(o^i_{t+n})_{1\le i \le k_{t+n}})))
\end{multline*}
where $k_{t+n}$ is the number of agents that are active at time $t+n$. Note, it is possible for $k_{t+n}$ to be greater or less than $k_t$ as any number of agents could have terminated early or been spawned at time step $t$. It is in this way that expected value from time $t+n$ may propagate to an agent that terminated at time $t$. 

\subsection{MA-POCA Counterfactual Baseline}\label{mapoca_baseline}
Agents who try to maximize a shared reward function suffer from a credit assignment problem since it is hard to disentangle an agent's actions contributed to the group's return.~\cite{Foerster18}. Counterfactual baselines (Eq.~\ref{counterfac}) marginalize out the action of an individual agent in the centralized state-action value function enabling the computation of a shaped, per-agent advantage. Though the architecture used in the original implementation of COMA has a number of advantages, it is limited to problems with discrete actions and a fixed number of agents. In this work, we propose an alternative leveraging self-attention, that alleviates both of these constraints.

We consider observations and observation-action pairs to be distinct entities. Letting $f_i: \mathcal{O}_i\times \mathcal{A}_i\rightarrow E$ be an encoding network for observation-action pairs, the counterfactual baseline can be learned explicitly by estimating the expectation in Eq.~\ref{counterfac} with Monte Carlo samples~\cite{Foerster18}. Thus, we can learn the counterfactual baseline for some agent $j$ by learning a value function that is conditioned on the observation-action pairs of all agents $i$ such that $\ 1\le i \le k_t\ i\ne j$ but \emph{only the observation} of agent $j$. Again using an $RSA$ block and observation and observation-action entity encoders, the baseline parameterized by $\psi$ for agent $j$ has the form
\begin{align*}
    Q_{\psi}(RSA(g_j(o_t^j), f_i(o^i_t, a^i_t)_{\substack{1\le i \le k_t\\ i\ne j}}))
\end{align*}
The objective for the baseline is
\begin{align}\label{baseline_objf}
    J(\psi) = (Q_{\psi}(RSA(g_j(o_t^j), f_i(o^i_t, a^i_t)_{\substack{1\le i \le k_t\\ i\ne j}})) - y^{(\lambda)})^2
\end{align}
which uses the \emph{same target} $y^{(\lambda)}$ as the value function update in Eq.~\ref{value_objf}. 

Note that a single joint observation $\boldsymbol{o} = (o^1,\ ...\ ,\ o^N)$ (and action) generates up to $N$ different samples on which to update Eq.~\ref{baseline_objf}, one for each $j$, $1 \le j \le N$. This is the key reason for using separate sets of parameters for the value function and baseline: in our training regime, the baseline is trained on the \emph{permutations} of all agent observations to estimate the \emph{per agent} baseline whereas the value function is not. This would mean a potential factor of $N$ more samples used to compute the baseline versus the value function. We hypothesize that this would lead to baseline dominance and, experimentally, we found using separate networks to perform better.

Finally, the advantage for agent $j$ to be used in the update in Eq.~\ref{policy_update} is given by
\begin{align*}\label{mapoca_adv}
     \text{Adv}_j = y^{(\lambda)} - Q_{\psi}(RSA(g_j(o_t^j), f_i(o^i_t, a^i_t)_{\substack{1\le i \le k_t\\ i\ne j}}))
\end{align*}

\section{Experiments}
In this section, we evaluate MA-POCA empirically on four multi-agent environments and compare its performance to the state-of-the-art multi-agent algorithm COMA~\cite{Foerster18} and the single-agent algorithm PPO~\cite{Schulman2017}. Three of the environments are built using Unity's ML-Agents Toolkit~\cite{Juliani2020}, and one is taken from the Multi-Agent Particle Environments~\cite{Lowe2017}. We choose to show the performance of PPO to illustrate that the environments require coordination to solve.

We show that, in standard cooperative tasks without dying or spawning, MA-POCA performs as well as or slightly better than COMA and that both outperform PPO. Furthermore, we show that MA-POCA significantly outperforms both baselines in tasks where agents die and/or spawn.  Note, we developed our own environments due to the lack of existing environments with these features. Code for all algorithms and environments is available.\footnote{https://github.com/Unity-Technologies/paper-ml-agents/tree/main/ma-poca}

\subsection{Algorithm and Baselines}

Architecturally, the implementation of COMA we use is similar to what is proposed in Section~\ref{method}. The key difference being the use of an $RSA$ block. Instead, all inputs are concatenated and fed into a fully connected neural network. In the case of agents that have terminated early or have not yet spawned, we use an absorbing state of all zeroes. 

In both MA-POCA and COMA, we use separate networks to approximate the value function and baseline as per the discussion in Section~\ref{mapoca_baseline}. Note, this is not a problem with the original COMA implementation~\cite{Foerster18} because their architecture depends on the assumption of strictly discrete actions which we do not make. However, the COMA architecture we use was suggested in the original work~\cite{Foerster18}.

We generate targets for the value function and baseline updates in Eqs.~\ref{value_objf} and \ref{baseline_objf} and analogues for COMA using $TD(\lambda)$~\cite{Sutton1988} as done in~\cite{Foerster18}. We use the value function and baseline network to compute the advantage for the policy update as in Eq.~\ref{mapocavf}. However, we do not use a target value function~\cite{mnih2015human} but instead use trust region clipping~\cite{Schulman2017} for the value function, baseline and policy updates which we found to work better in practice.

We use the implementation of Proximal Policy Optimization (PPO)~\cite{Schulman2017}, and generalized advantage estimation (GAE)~\cite{Schulman16} contained in the Unity ML-Agents Toolkit~\cite{Juliani2020}.

\begin{figure*}[ht!]
\begin{center}
    \setlength{\tabcolsep}{1pt}
    \begin{tabular*}{\textwidth}[t]{@{\extracolsep{\fill}}cccc}
        \includegraphics[width=.24\textwidth]{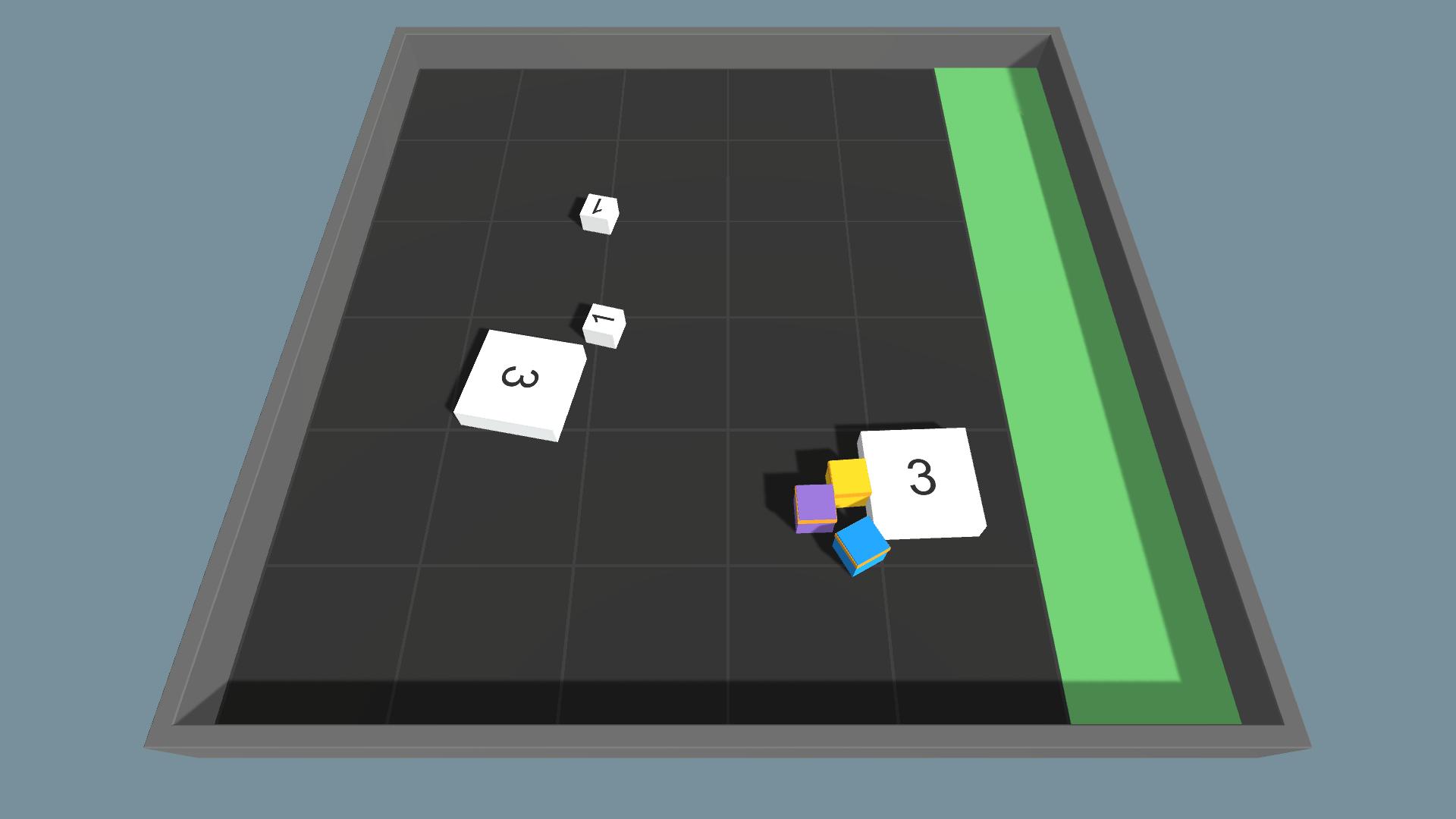}&
        \includegraphics[width=.24\textwidth]{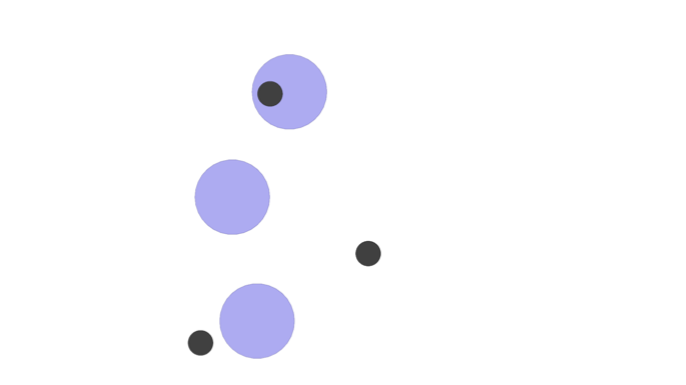}& 
        \includegraphics[width=.24\textwidth]{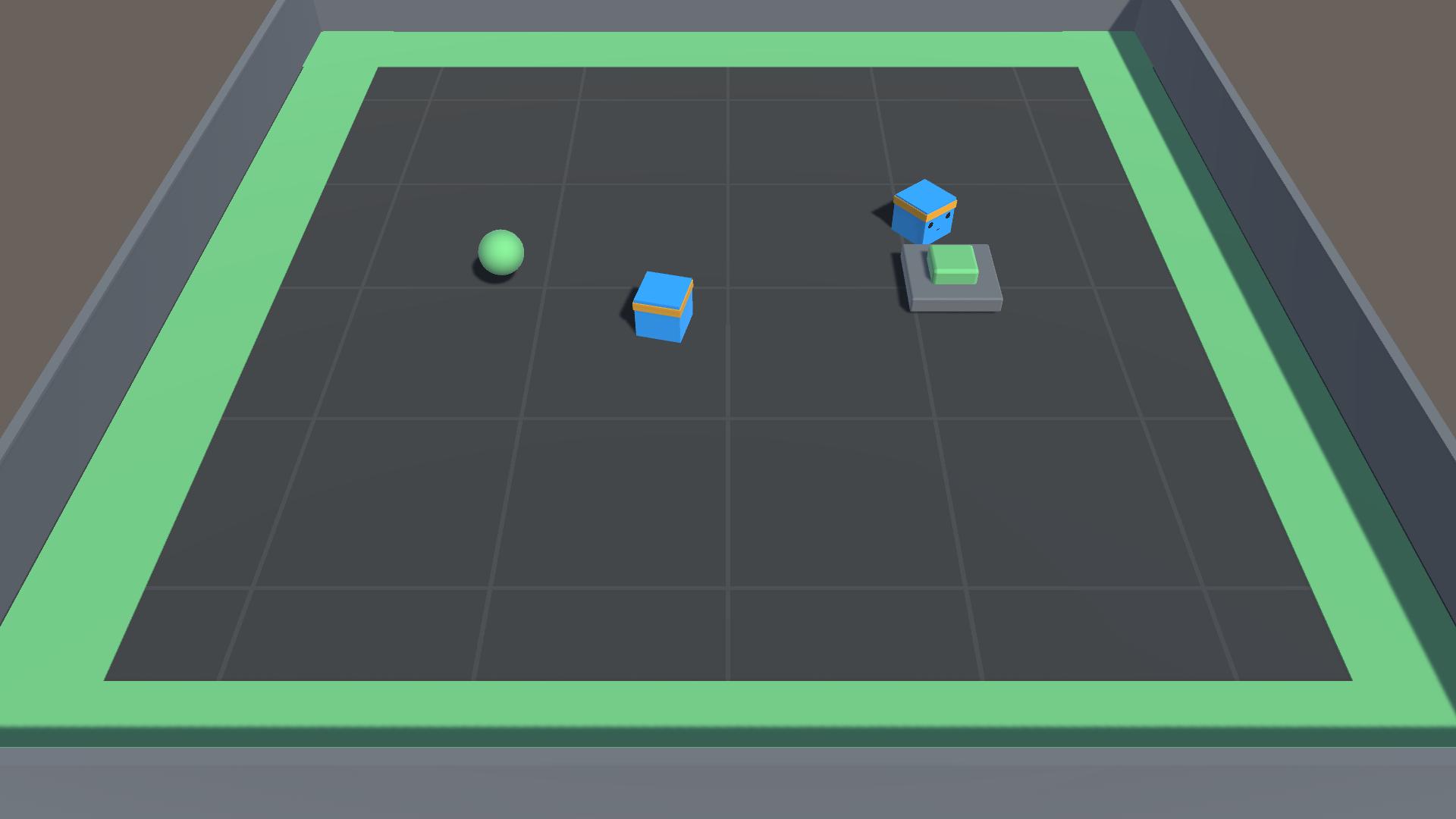}& 
        \includegraphics[width=.24\textwidth]{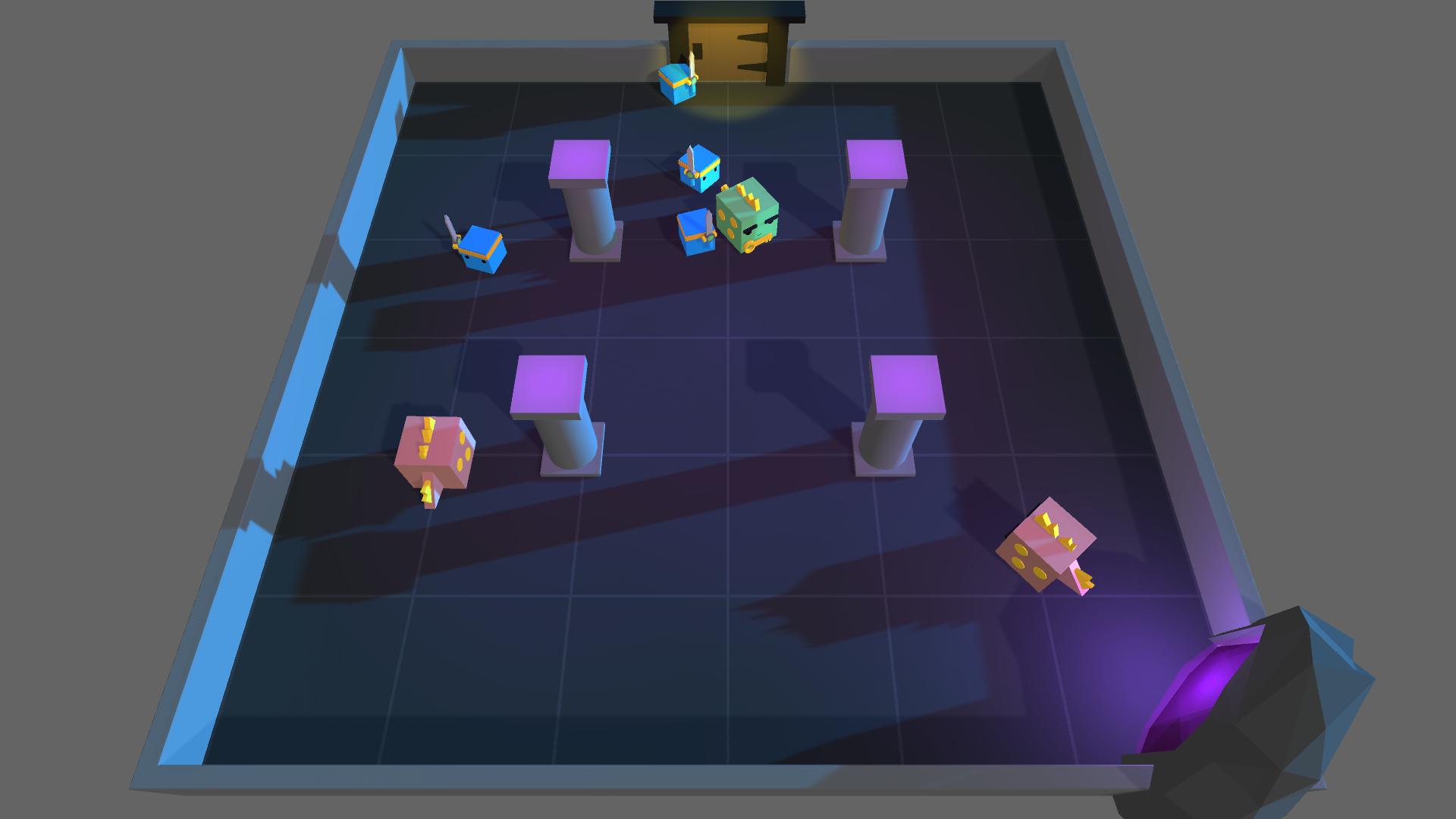}\\
        (a) & (b) & (c) & (d)
        \end{tabular*}
    \end{center}
    \caption{\textbf{(a) Collaborative Push Block}. Agents (blue, yellow, purple) must push white blocks to green area; larger blocks require more agents to push. \textbf{(b) Simple Spread}. Agents (large circles) must move to cover targets (small circles) without colliding with one another. \textbf{(c) Baton Pass}. Blue agents must grab green food and hit green button to spawn another agent, who can grab the next food, and so on. \textbf{(d) Dungeon Escape}. Blue agents must kill green dragon by sacrificing one of them to reveal a key. Teammates must pick up key and reach the door, while avoiding pink dragons.}
    \label{fig:environments}
\end{figure*}

\subsection{Results}
A brief description of the environments is provided in Figure~\ref{fig:environments}; further details can be found in Appendix~\ref{appendix:environments}. Figure~\ref{fig:results} compares the mean and $95\%$ confidence interval of episodic reward for MA-POCA, COMA, and PPO over 10 seeds each. Hyperparameters for all algorithms and experiments are contained in Appendix~\ref{sec:hyperparameters}.  

In all four environments, PPO is unable to find the optimal policies and converges to a local optima. This is likely due to the partial observability introduced by exclusively decentralized training and acting. In (c) and (d), PPO has no mechanism to address the posthumous credit assignment problem so it is unable to learn from value beyond it's termination. For example, in (d) Dungeon Escape, the PPO agents are able to use the key when they observe it--however, they are not able to learn that killing the green dragon is the way to get the key as this removes them from the environment. Thus, they solve the problem roughly half the time when they accidentally collide with the green dragon. Also, note that, the reward is decreasing indicating that they are learning to avoid the green dragon (possibly, to hold out for the key when it drops).

The curves in (a) and (b) in the top row of Figure~\ref{fig:results} contain results for the Collaborative Push Block and Simple Spreader environments which do not contain spawning or dying agents. In these environments, MA-POCA learns slightly faster than COMA. We hypothesize that the permutation invariance of attention gives MA-POCA an advantage as COMA's value network needs to learn that any permutation of a joint observation has the same value. Additionally, the cross-comparison of entities in attention may enable more robust modeling of the group value function.

The curves in (c) and (d) in the bottom row of Figure~\ref{fig:results} contain results for the Baton Pass and Dungeon Escape environments which do contain spawning and/or dying agents. In these environments, MA-POCA significantly outperforms COMA with less variance between seeds. COMA eventually converges to the optimal policy. Of particular interest is that both PPO and MA-POCA are faster initially than COMA. We hypothesize COMA's inferior sample complexity is due to the inefficient input representation that absorbing states provide, as discussed in Section~\ref{sec:abs_states}.

\begin{figure*}[ht!]
    \setlength{\tabcolsep}{1pt}
    \begin{tabular}[t]{cccc}
        \multicolumn{2}{c}{
        \includegraphics[width=.47\textwidth]{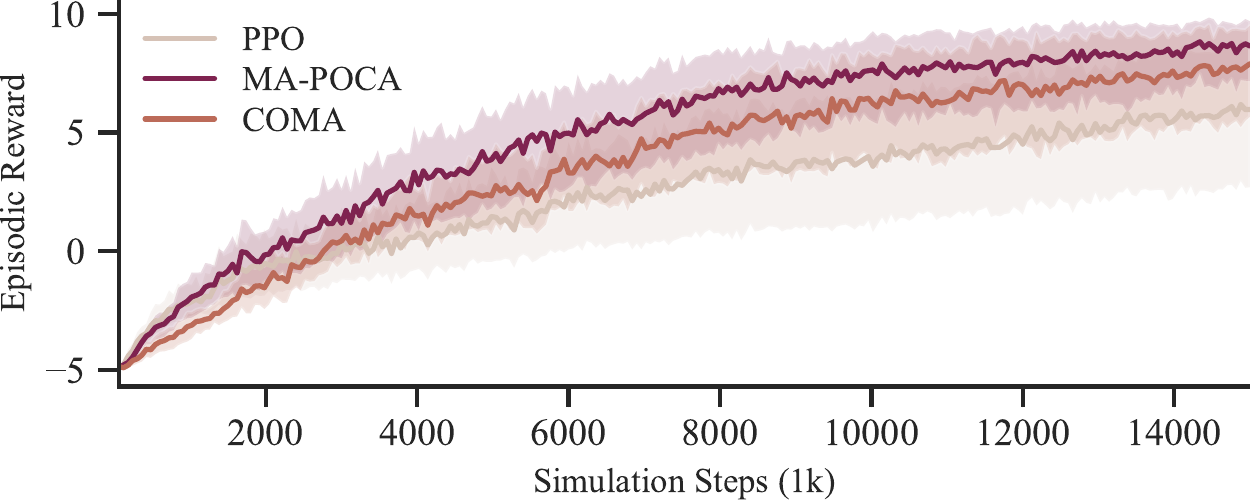}} & \multicolumn{2}{c}{\includegraphics[width=.47\textwidth]{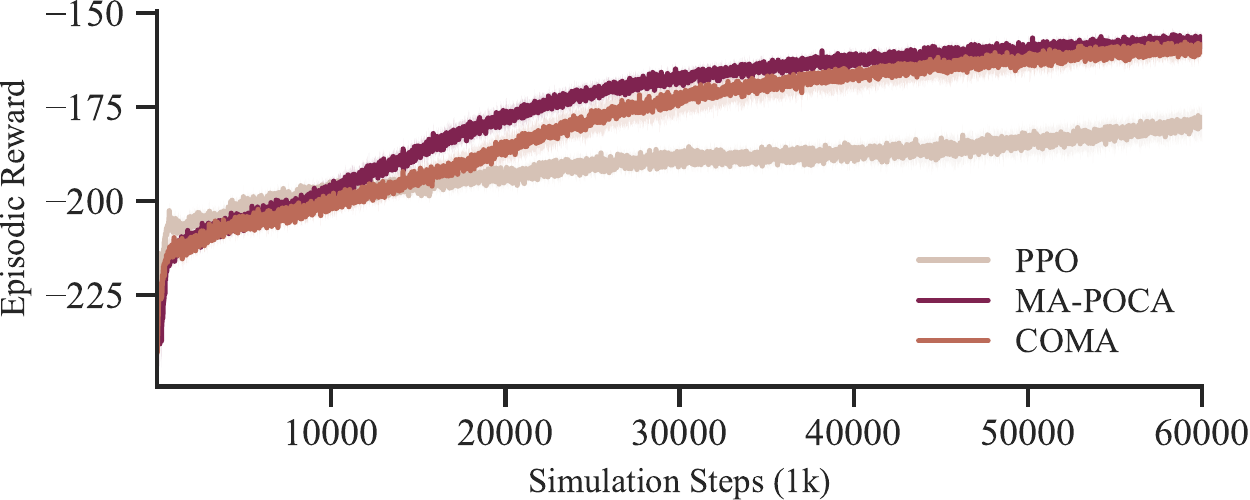}
        } \\
        \multicolumn{2}{c}{(a) Collaborative Push Block} & \multicolumn{2}{c}{(b) Simple Spread} \\ 
        \multicolumn{2}{c}{
        \includegraphics[width=.47\textwidth]{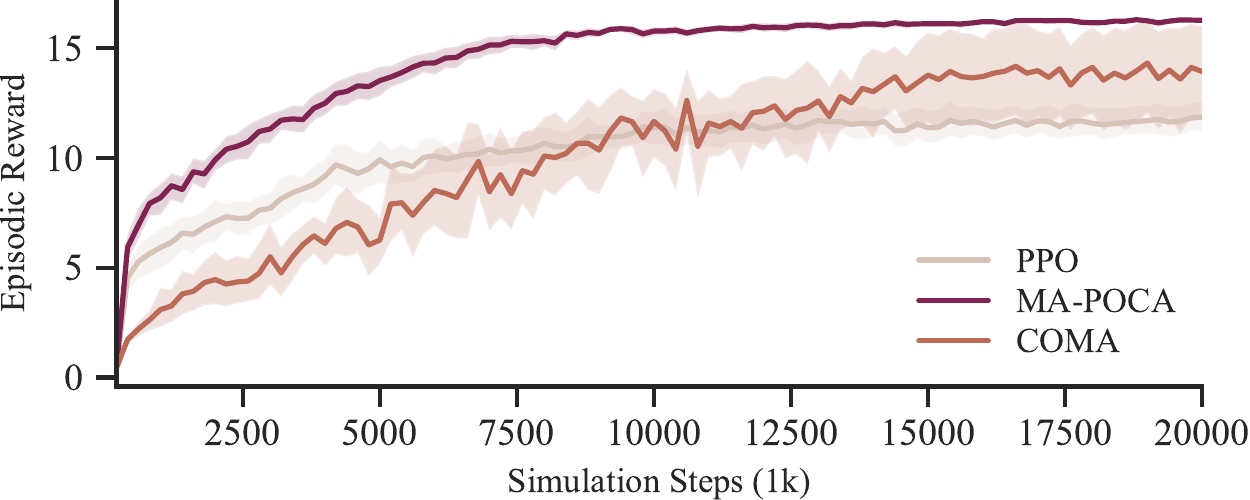}} & \multicolumn{2}{c}{\includegraphics[width=.47\textwidth]{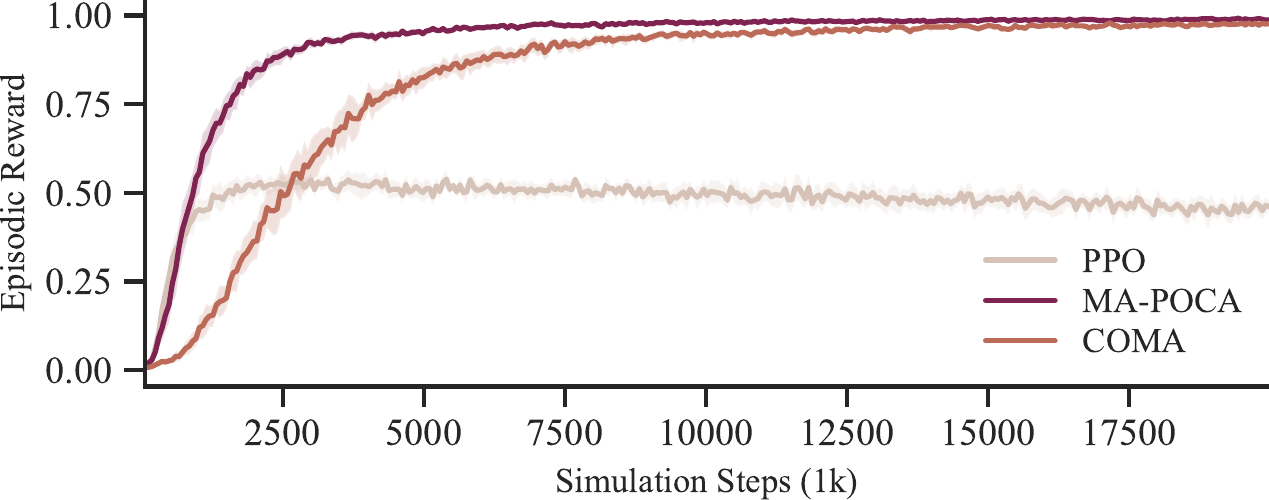}
        }\\
        \multicolumn{2}{c}{(c) Baton Pass} & \multicolumn{2}{c}{(d) Dungeon Escape} \\
        \end{tabular}
   \caption{Comparison between MA-POCA, COMA, and PPO of cumulative reward per episode for {\bf (a)} Collaborative Push Block, {\bf (b)} Simple Spread, {\bf (c)} Baton Pass, {\bf (d)} Dungeon Escape. Results are averaged across 10 seeds. MA-POCA outperforms COMA in all four environments, and significantly so in environments that involve agents spawning or dying ({\bf (c)} and {\bf (d)}). PPO converges to a sub-optimal policy in all tasks.}\label{fig:results}
\end{figure*}
\section{Related Work}
\subsubsection{Group-Centric MARL}
Value decomposition methods~\cite{Sunehag17,Rashid18,Iqbal21} are an alternative to the counterfactual baseline used by COMA and MA-POCA to address multi-agent credit assignment. Value decomposition makes the assumption that the group value function is a monotonic function of the individual's value functions and thus assumes actions are independent which is not true in general.

\subsubsection{Varying Agents in MARL}
Actor-Attention-Critic~\cite{pmlr-v97-iqbal19a} uses an attention-based state-action value function for each agent to estimate an individual's expected reward conditioned on the state of other agents, though the number of agents remains constant per episode. Similarly, Graph Policy Learning~\cite{pmlr-v139-rahman21a} uses a graph neural network to condition an agent's value function and policy on the state of a varying number of teammates. However, as both approaches depend on the individual agent to condition the value function, they do not address posthumous rewards without absorbing states. Evolutionary Population Curriculum~\cite{Long20} uses attention to handle a population of agents increasing in size but also uses distinct value function for each agent and so would also require absorbing states for posthumous credit assignment. Randomized Entity-Wise Factorization~\cite{Iqbal21} applies attention dynamically group agents, but does not address a variable number of total agents.

\subsubsection{Environment Implementations}
The commonly used StarCraft Multi-Agent Challenge (SMAC)~\cite{samvelyan19smac} benchmarks return all zeros as an absorbing state for units that have died. Death masking~\cite{Yu2021} is a variant of absorbing state which appends an agent ID to a vector of all zeros. This is an instance of using different absorbing states for different outcomes as discussed in Section~\ref{sec:abs_states}.

\subsubsection{Support for Varying Agents}
To the author's knowledge, there are three APIs with explicit support for dying and spawning agents: PettingZoo~\cite{terry2020pettingzoo}, RLLib~\cite{Liang2018}, and the Unity ML-Agents Toolkit~\cite{Juliani2020}. RLLib also contains implementations of MARL algorithms though all would require absorbing states to be used with varying numbers of agents.

\section{Conclusion}
This paper explicitly identified the Posthumous Credit Assignment problem created when agents terminate early. The is currently handled in MARL by adding an absorbing state for agents that terminate early. Using a toy supervised learning problem, we empirically demonstrated a downside of using absorbing states. We then introduced MA-POCA, a novel architecture designed to train groups of agents to solve tasks in which individual agents may terminate early or spawn, without the use of absorbing states. MA-POCA naturally handles varying numbers of agents and achieves a counterfactual baseline via the use of self-attention. Finally, we demonstrate that MA-POCA outperforms COMA and PPO on two standard MARL tasks without dying or spawning agents. More importantly, MA-POCA significantly outperforms both on tasks with dying and spawning agents. Future work will extend other algorithms in the decentralized POMDP framework beyond absorbing states and investigate potential formalisms for problems where the maximum number of agents $N$ is unknown.


\pagebreak
\bibliography{Cohen_etal}

\begin{thebibliography}{27}
\providecommand{\natexlab}[1]{#1}

\bibitem[{Ba, Kiros, and Hinton(2016)}]{ba2016layer}
Ba, J.~L.; Kiros, J.~R.; and Hinton, G.~E. 2016.
\newblock Layer Normalization.
\newblock arXiv:1607.06450.

\bibitem[{Baker et~al.(2020)Baker, Kanitscheider, Markov, Wu, Powell, McGrew,
  and Mordatch}]{Baker20}
Baker, B.; Kanitscheider, I.; Markov, T.; Wu, Y.; Powell, G.; McGrew, B.; and
  Mordatch, I. 2020.
\newblock Emergent Tool Use From Multi-Agent Autocurricula.
\newblock In \emph{International Conference on Learning Representations}.

\bibitem[{Espeholt et~al.(2018)Espeholt, Soyer, Munos, Simonyan, Mnih, Ward,
  Yotam, Vlad, Tim, Dunning, Legg, and Kavukcuoglu}]{Espeholt2018}
Espeholt, L.; Soyer, H.; Munos, R.; Simonyan, K.; Mnih, V.; Ward, T.; Yotam,
  B.; Vlad, F.; Tim, H.; Dunning, I.; Legg, S.; and Kavukcuoglu, K. 2018.
\newblock {IMPALA: Scalable Distributed Deep-RL with Importance Weighted
  Actor-Learner Architectures}.
\newblock In \emph{International Conference on Machine Learning}, volume~4,
  2263--2284.
\newblock ISBN 9781510867963.

\bibitem[{Foerster et~al.(2018)Foerster, Farquhar, Afouras, Nardelli, and
  Whiteson}]{Foerster18}
Foerster, J.; Farquhar, G.; Afouras, T.; Nardelli, N.; and Whiteson, S. 2018.
\newblock Counterfactual Multi-Agent Policy Gradients.
\newblock In \emph{Proceedings of the Thirty-Second AAAI Conference on
  Artificial Intelligence}.

\bibitem[{Horgan et~al.(2018)Horgan, Quan, Budden, Barth-maron, Hessel, van
  Hasselt, and Silver}]{Horgan2018}
Horgan, D.; Quan, J.; Budden, D.; Barth-maron, G.; Hessel, M.; van Hasselt, H.;
  and Silver, D. 2018.
\newblock {Distributed Prioritized Experience Replay}.
\newblock In \emph{International Conference on Learning Representations},
  1--19.

\bibitem[{Iqbal et~al.(2021)Iqbal, Schroeder~de Witt, Peng, B{\"o}hmer,
  Whiteson, and Sha}]{Iqbal21}
Iqbal, S.; Schroeder~de Witt, C.; Peng, B.; B{\"o}hmer, W.; Whiteson, S.; and
  Sha, F. 2021.
\newblock Randomized Entity-Wise Factorization for Multi-Agent Reinforcement
  Learning.
\newblock In \emph{Proceedings of the 38th International Conference on Machine
  Learning}.

\bibitem[{Iqbal and Sha(2019)}]{pmlr-v97-iqbal19a}
Iqbal, S.; and Sha, F. 2019.
\newblock Actor-Attention-Critic for Multi-Agent Reinforcement Learning.
\newblock In Chaudhuri, K.; and Salakhutdinov, R., eds., \emph{Proceedings of
  the 36th International Conference on Machine Learning}, volume~97 of
  \emph{Proceedings of Machine Learning Research}, 2961--2970. Long Beach,
  California, USA: PMLR.

\bibitem[{Juliani et~al.(2020)Juliani, Berges, Teng, Cohen, Harper, Elion, Goy,
  Gao, Henry, Mattar, and Lange}]{Juliani2020}
Juliani, A.; Berges, V.; Teng, E.; Cohen, A.; Harper, J.; Elion, C.; Goy, C.;
  Gao, Y.; Henry, H.; Mattar, M.; and Lange, D. 2020.
\newblock Unity: {A} General Platform for Intelligent Agents.
\newblock \emph{arXiv preprint}, abs/1809.02627.

\bibitem[{Konda and Tsitsiklis(2000)}]{Konda00}
Konda, V.; and Tsitsiklis, J. 2000.
\newblock Actor-critic algorithms.
\newblock In \emph{Advances in Neural Information Processing Systems}.

\bibitem[{Liang et~al.(2018)Liang, Liaw, Moritz, Nishihara, Fox, Goldberg,
  Gonzalez, Jordan, and Stoica}]{Liang2018}
Liang, E.; Liaw, R.; Moritz, P.; Nishihara, R.; Fox, R.; Goldberg, K.;
  Gonzalez, J.~E.; Jordan, M.~I.; and Stoica, I. 2018.
\newblock RLlib: Abstractions for Distributed Reinforcement Learning.
\newblock \emph{arXiv preprint arXiv:1712.09381}.

\bibitem[{Littman(1994)}]{Littman94}
Littman, M. 1994.
\newblock Markov-games as a framework for multi-agent reinforcement learning.
\newblock In \emph{International Conference on Machine Learning}, 157--163.

\bibitem[{Long et~al.(2020)Long, Zhou, Gupta, Fang, Wu, and Wang}]{Long20}
Long, Q.; Zhou, Z.; Gupta, A.; Fang, F.; Wu, Y.; and Wang, X. 2020.
\newblock Evolutionary Population Curriculum for Scaling Multi-Agent
  Reinforcement Learning.
\newblock In \emph{International Conference on Learning Representations}.

\bibitem[{Lowe et~al.(2017)Lowe, Wu, Tamar, Harb, Abbeel, and
  Mordatch}]{Lowe2017}
Lowe, R.; Wu, Y.; Tamar, A.; Harb, J.; Abbeel, P.; and Mordatch, I. 2017.
\newblock Multi-Agent Actor-Critic for Mixed Cooperative-Competitive
  Environments.
\newblock \emph{Neural Information Processing Systems (NIPS)}.

\bibitem[{Lyu et~al.(2021)Lyu, Xiao, Daley, and Amato}]{Lyu21}
Lyu, X.; Xiao, Y.; Daley, B.; and Amato, C. 2021.
\newblock Contrasting Centralized and Decentralized Critics in Mult-Agent
  Reinforcement Learning.
\newblock \emph{arXiv preprint arXiv: 2102.04402}.

\bibitem[{Mnih et~al.(2015)Mnih, Kavukcuoglu, Silver, Rusu, Veness, Bellemare,
  Graves, Riedmiller, Fidjeland, Ostrovski, Petersen, Beattie, Sadik,
  Antonoglou, King, Kumaran, Wierstra, Legg, and Hassabis}]{mnih2015human}
Mnih, V.; Kavukcuoglu, K.; Silver, D.; Rusu, A.~A.; Veness, J.; Bellemare,
  M.~G.; Graves, A.; Riedmiller, M.; Fidjeland, A.~K.; Ostrovski, G.; Petersen,
  S.; Beattie, C.; Sadik, A.; Antonoglou, I.; King, H.; Kumaran, D.; Wierstra,
  D.; Legg, S.; and Hassabis, D. 2015.
\newblock Human-level control through deep reinforcement learning.
\newblock \emph{Nature}, 518(7540): 529.

\bibitem[{Oliehoek and Amato(2016)}]{Oliehoek}
Oliehoek, F.~A.; and Amato, C. 2016.
\newblock \emph{A Concise Introduction to Decentralized POMDPs}.
\newblock Springer.

\bibitem[{Rahman et~al.(2021)Rahman, Hopner, Christianos, and
  Albrecht}]{pmlr-v139-rahman21a}
Rahman, M.~A.; Hopner, N.; Christianos, F.; and Albrecht, S.~V. 2021.
\newblock Towards Open Ad Hoc Teamwork Using Graph-based Policy Learning.
\newblock In Meila, M.; and Zhang, T., eds., \emph{Proceedings of the 38th
  International Conference on Machine Learning}, volume 139 of
  \emph{Proceedings of Machine Learning Research}, 8776--8786. PMLR.

\bibitem[{Rashid et~al.(2018)Rashid, Samvelyan, Schroeder~de Witt, Farquhar,
  Foerster, and Whiteson}]{Rashid18}
Rashid, T.; Samvelyan, M.; Schroeder~de Witt, C.; Farquhar, G.; Foerster, J.;
  and Whiteson, S. 2018.
\newblock QMIX: Monotonic Value Function Factprisation for Deep Multi-Agent
  Reinforcement Learning.
\newblock In \emph{Proceedings of the 35th International Conference on Machine
  Learning}.

\bibitem[{Samvelyan et~al.(2019)Samvelyan, Rashid, de~Witt, Farquhar, Nardelli,
  Rudner, Hung, Torr, Foerster, and Whiteson}]{samvelyan19smac}
Samvelyan, M.; Rashid, T.; de~Witt, C.~S.; Farquhar, G.; Nardelli, N.; Rudner,
  T. G.~J.; Hung, C.-M.; Torr, P. H.~S.; Foerster, J.; and Whiteson, S. 2019.
\newblock {The} {StarCraft} {Multi}-{Agent} {Challenge}.
\newblock \emph{CoRR}, abs/1902.04043.

\bibitem[{Schulman et~al.(2016)Schulman, Moritz, Levine, Jordan, and
  Abbeel}]{Schulman16}
Schulman, J.; Moritz, P.; Levine, S.; Jordan, M.~I.; and Abbeel, P. 2016.
\newblock High-Dimensional Continuous Control Using Generalzied Advantage
  Estimation.
\newblock In \emph{International Conference on Learning Representations}.

\bibitem[{Schulman et~al.(2017)Schulman, Wolski, Dhariwal, Radford, and
  Klimov}]{Schulman2017}
Schulman, J.; Wolski, F.; Dhariwal, P.; Radford, A.; and Klimov, O. 2017.
\newblock Proximal Policy Optimization Algorithms.
\newblock \emph{arXiv preprint}, abs/1707.06347.

\bibitem[{Sunehag et~al.(2017)Sunehag, Lever, Gruslys, Marian~Czarnecki,
  Zambaldi, Jaderberg, Lanctot, Sonnerat, Z.~Leibo, Tuyls, and
  Graepel}]{Sunehag17}
Sunehag, P.; Lever, G.; Gruslys, A.; Marian~Czarnecki, W.; Zambaldi, V.;
  Jaderberg, M.; Lanctot, M.; Sonnerat, N.; Z.~Leibo, J.; Tuyls, K.; and
  Graepel, T. 2017.
\newblock Value-Decomposition Networks For Cooperative Multi-Agent Learning.
\newblock In \emph{Proceedings of the 17th International Conference on
  Autonomous Agents and Multiagent Systems}.

\bibitem[{Sutton(1988)}]{Sutton1988}
Sutton, R.~S. 1988.
\newblock Learning to Predict by the Methods of Temporal Differences.
\newblock \emph{Machine Learning}, 3(1): 9–44.

\bibitem[{Terry et~al.(2020)Terry, Black, Grammel, Jayakumar, Hari, Sulivan,
  Santos, Perez, Horsch, Dieffendahl, Williams, Lokesh, Sullivan, and
  Ravi}]{terry2020pettingzoo}
Terry, J.~K.; Black, B.; Grammel, N.; Jayakumar, M.; Hari, A.; Sulivan, R.;
  Santos, L.; Perez, R.; Horsch, C.; Dieffendahl, C.; Williams, N.~L.; Lokesh,
  Y.; Sullivan, R.; and Ravi, P. 2020.
\newblock PettingZoo: Gym for Multi-Agent Reinforcement Learning.
\newblock \emph{arXiv preprint arXiv:2009.14471}.

\bibitem[{Vaswani et~al.(2017)Vaswani, Shazeer, Parmar, Uszkoreit, Jones,
  Gomez, Kaiser, and Polosukhin}]{Vaswani2017}
Vaswani, A.; Shazeer, N.; Parmar, N.; Uszkoreit, J.; Jones, L.; Gomez, A.~N.;
  Kaiser, {\L}.; and Polosukhin, I. 2017.
\newblock {Attention is All You Need}.
\newblock In \emph{Advances in Neural Information Processing Systems},
  5999--6009.

\bibitem[{Wolpert and Tumer(2002)}]{Wolpert02}
Wolpert, D.; and Tumer, K. 2002.
\newblock Optimal Payoff Functions for Members of Collectives.
\newblock In \emph{Modeling Complexity in Economic and Social Systems},
  355--369. World Scientific.

\bibitem[{Yu et~al.(2021)Yu, Velu, Vinitsky, Wang, and Bayen}]{Yu2021}
Yu, C.; Velu, A.; Vinitsky, E.; Wang, Y.; and Bayen, A. 2021.
\newblock The Surprising Effectiveness of PPO in Cooperative Multi-Agent Games.
\newblock \emph{arXiv preprint arXiv:2103.01955}.

\end{thebibliography}
\pagebreak
\appendix
\section{Additional Experiments for Mean Computation}\label{abs_curves}
\begin{figure*}[ht!]
    \begin{tabular}{ccc}
    \includegraphics[width=.33\textwidth]{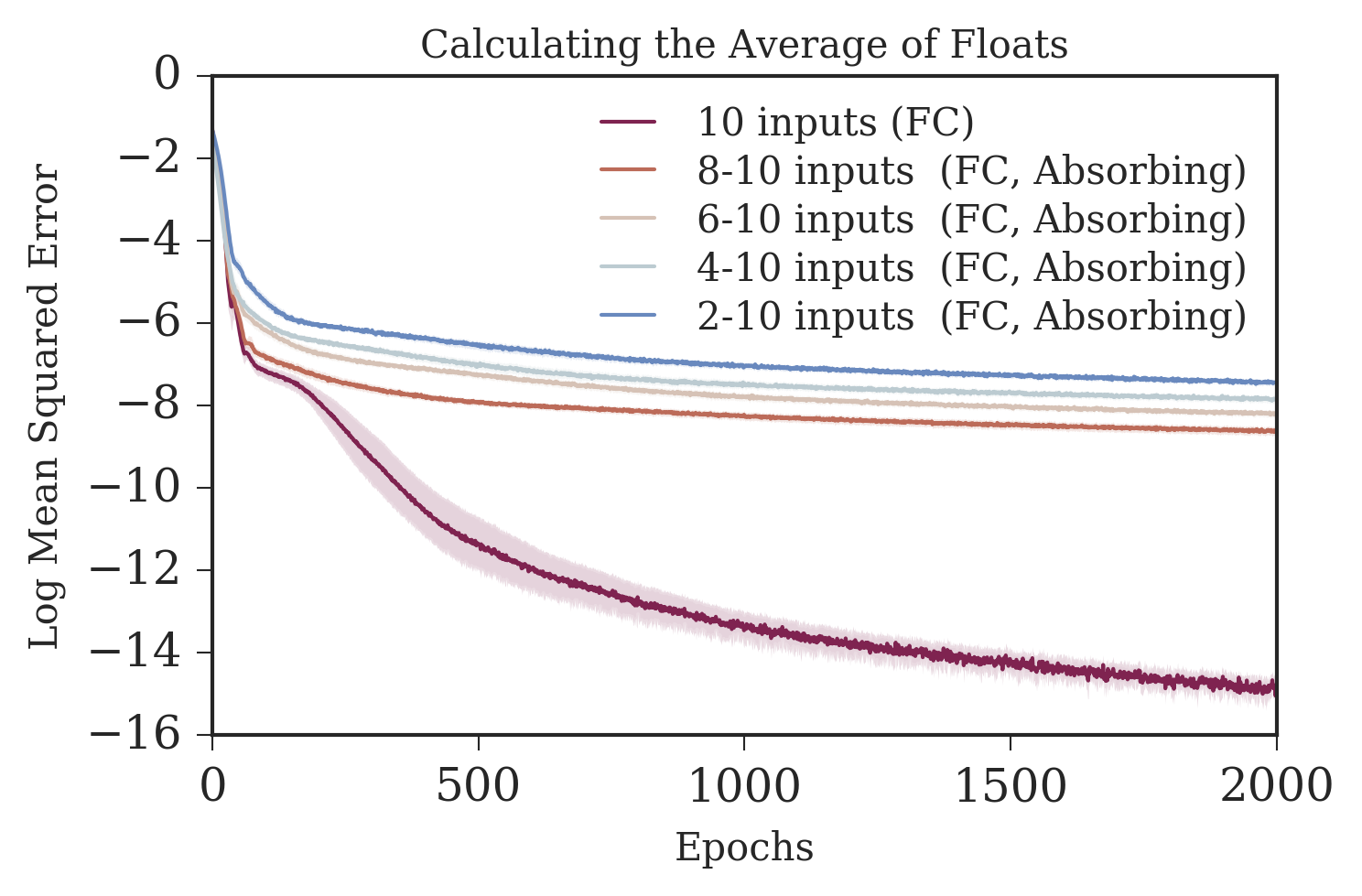} & \includegraphics[width=.33\textwidth]{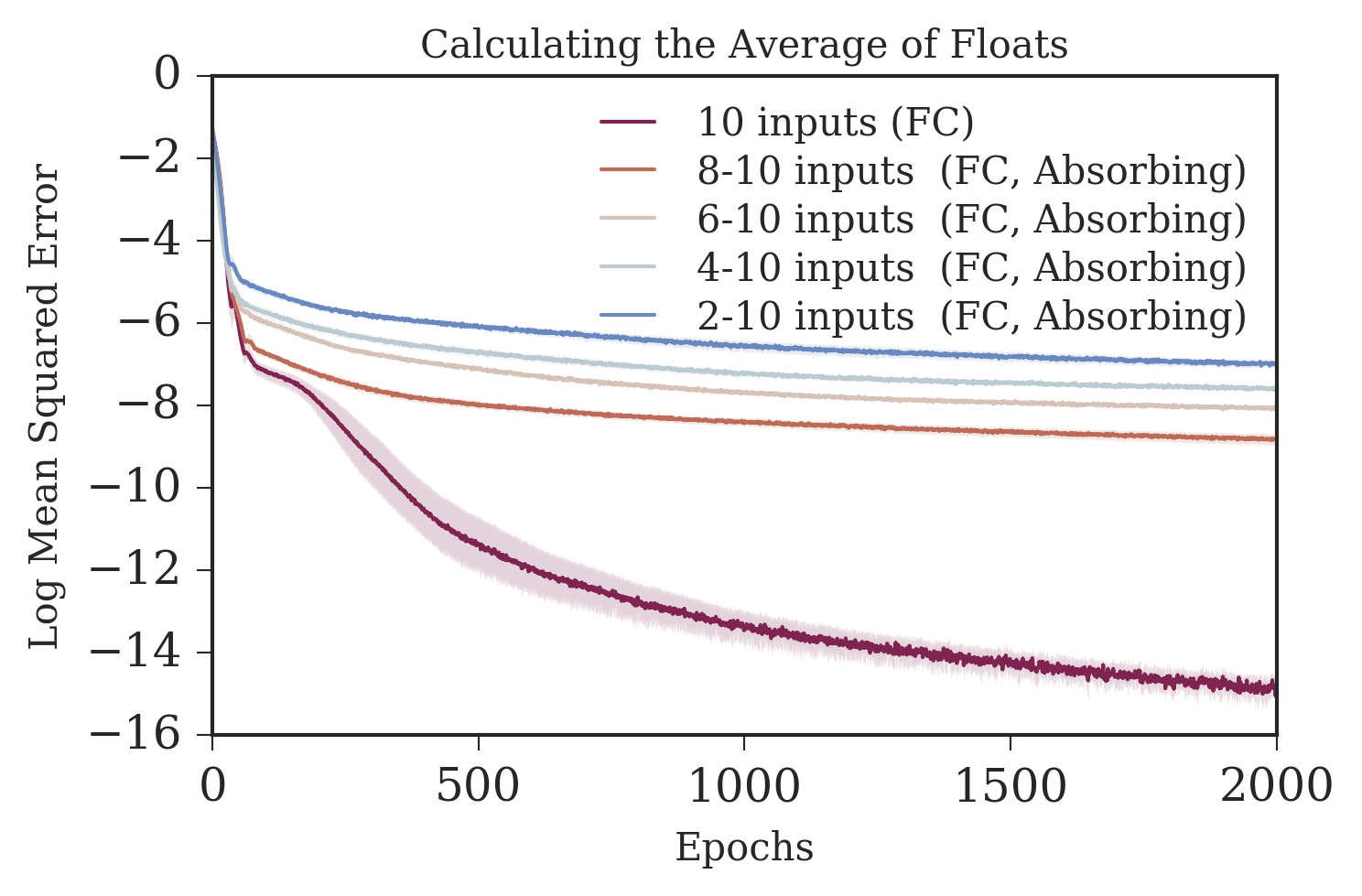} & \includegraphics[width=.33\textwidth]{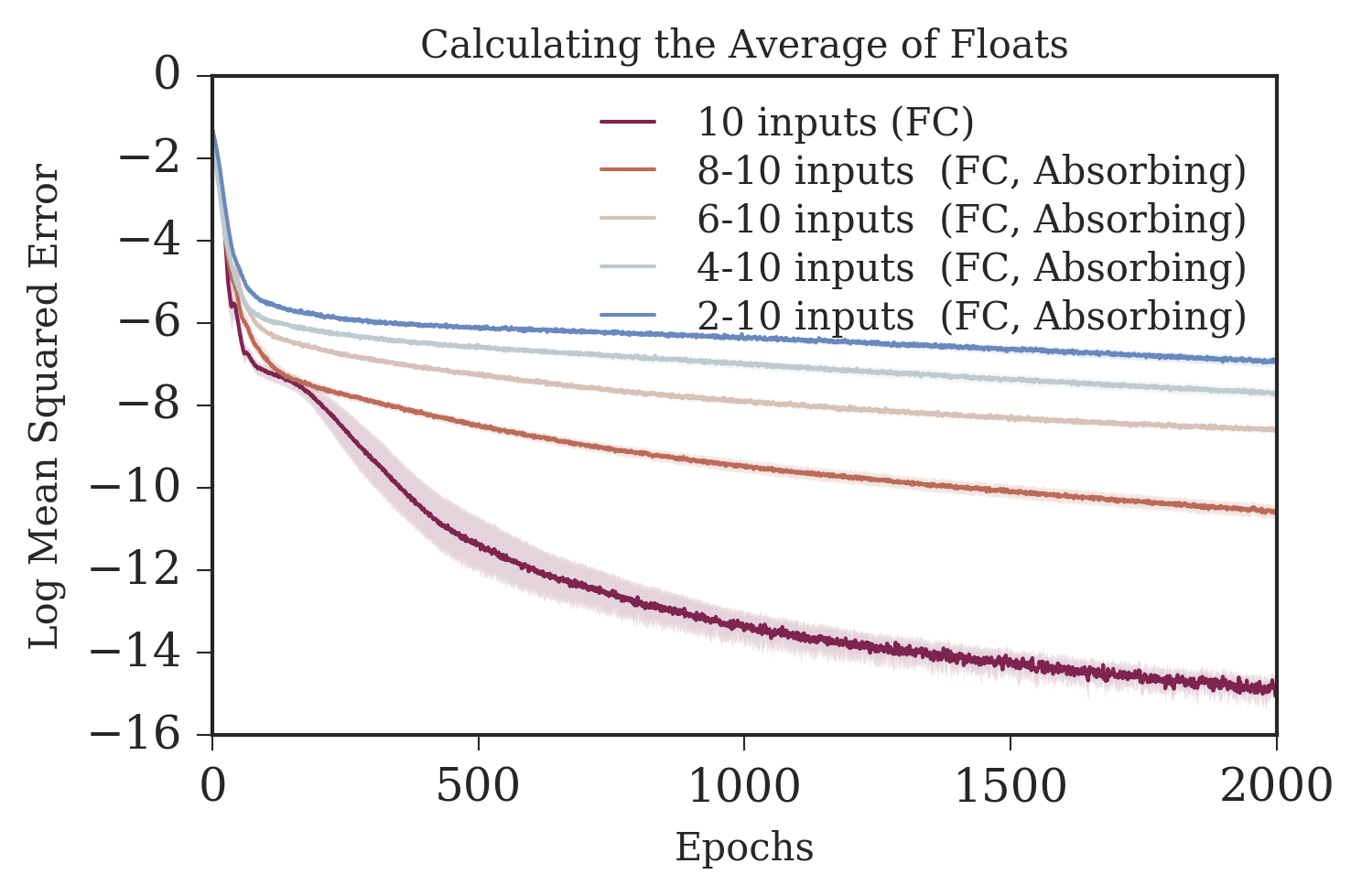} \\
    $o_{abs} = 0$ & $o_{abs} = 1$ & $o_{abs} = -1$
    \end{tabular}
    \caption{The sample efficiency of learning to compute the mean of a varying number of floats depends on the quantity of absorbing states. The performance for $o_{abs} = 0.0\, -1.0,\ 1.0$ are roughly the same and asymptotically all converge. Finally, without absorbing states (10 inputs (FC)) the task is trivial.}\label{abs0}
\end{figure*}

\begin{figure}[H]
    \includegraphics[width=.5\textwidth]{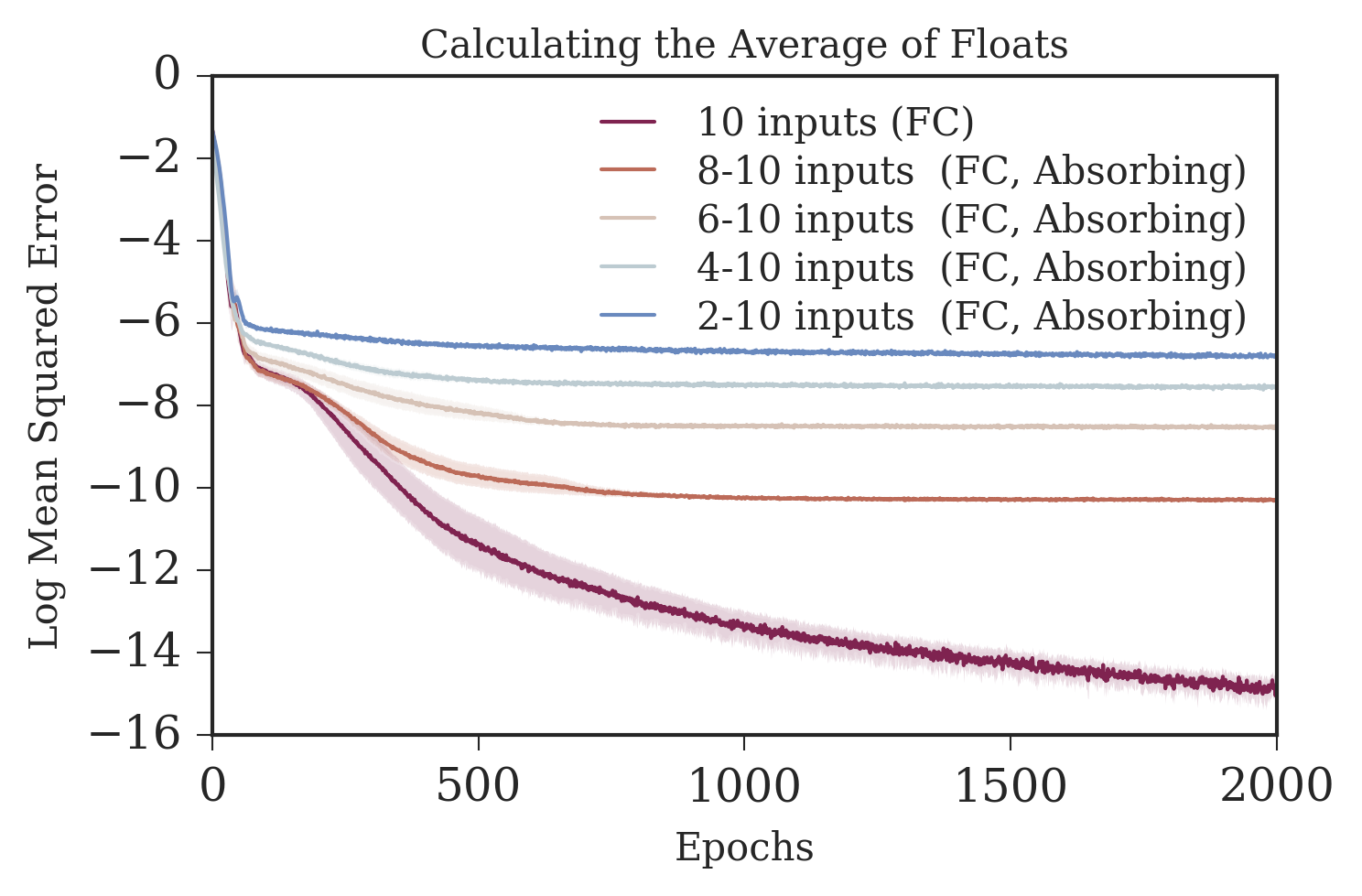}
    \caption{$o_{abs} = 0.4$. Since $o_{abs} \in [0.25, 0.75]$, this problem is partially observable and asymptotically the network cannot converge to solve the task. }\label{bad_abs}
\end{figure}

\begin{figure}[H]
    \includegraphics[width=.5\textwidth]{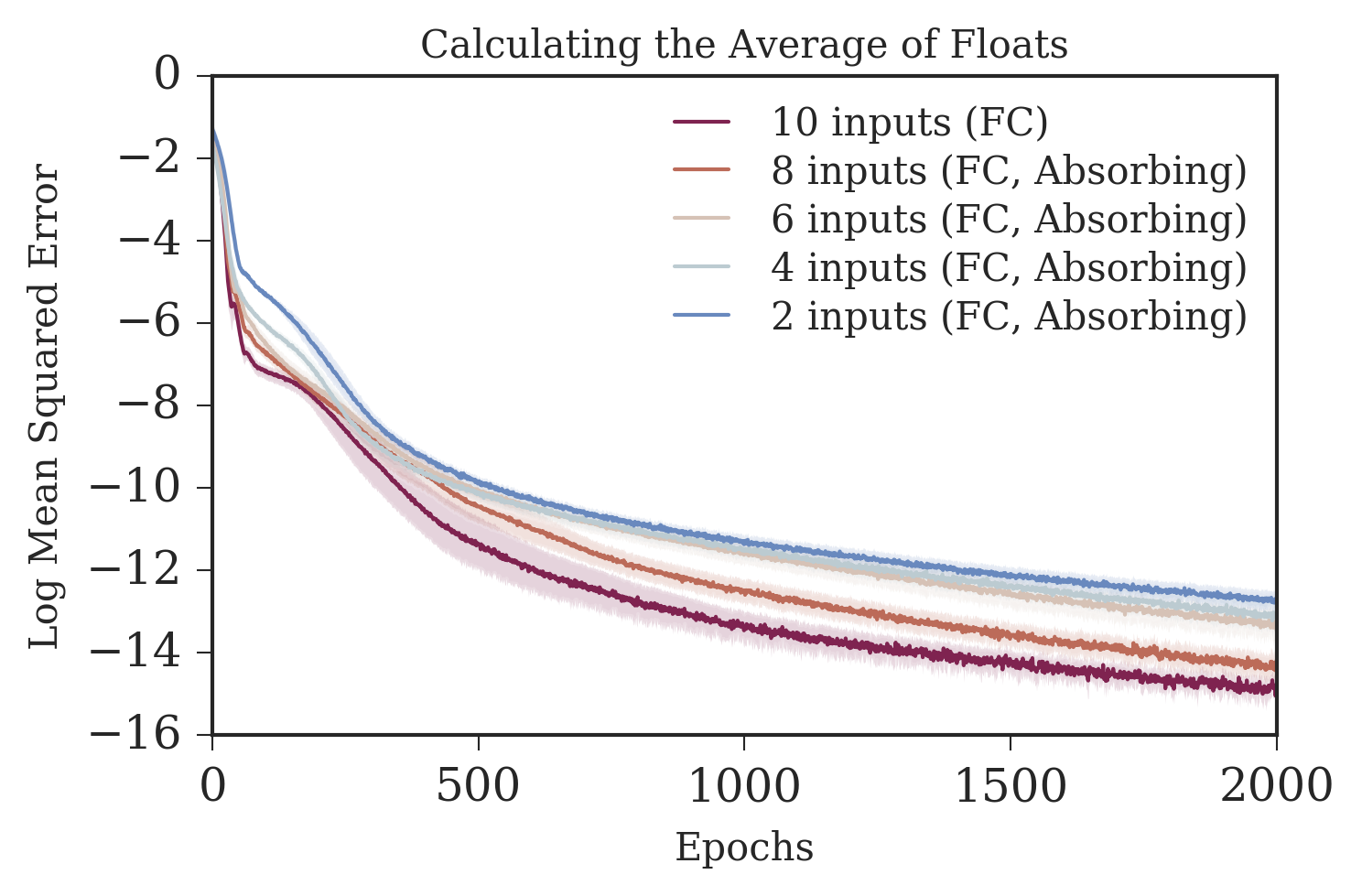}
    \caption{The number of absorbing states per sample is fixed but still shuffled.  This version of the task is significantly simpler than when the number of absorbing states is varied per sample.}\label{fixed_abs}
\end{figure}

In this section, we provide additional figures for the different values of $o_{abs} = -1.0,\ 1.0,\ 0.4$ showing that $-1.0$ and $1.0$ are reasonable choices (see Figure~\ref{abs0}) but $0.4$ is not as it is contained in $[0.25, 0.75]$ (see Figure~\ref{bad_abs}). Additionally, we provide a figure for the curves when we fix the number of absorbing states (see Figure~\ref{fixed_abs}) to demonstrate that varying the number of floats per sample is more challenging. All hyperparameters are contained in Appendix~\ref{sec:hyperparameters} and are the same as the experiment discussed in the main text.  In each figure, curves are the mean and $95\%$ confidence interval over 20 seeds.

\section{Self-Attention}\label{atten_app}



MA-POCA's centralized critic uses an $RSA$ module in order to process a variable number of agents. The agent's observations are first embedded using a fully connected layer~\cite{Baker20}. Each agent's observation embedding is normalized using Layer Normalization~\cite{ba2016layer} and then further embedded into Query : $Q$, Key : $K$ and Value: $V$ using a fully connected network. $Q$, $K$ and $V$ are fed into a scaled dot-product multi-head attention~\cite{Vaswani2017}. The original observation embeddings are summed with the processed embeddings (the residual connection) and normalized again with Layer Normalization. The resulting embeddings are then averaged together to form a fixed size embedding. The same attention mechanism is used in the calculation of average experiments shown in Figure \ref{fig:abs_ablate}.

\section{Environments}
\label{appendix:environments}

\textbf{Cooperative Push Block.} In this environment, three agents must push 5 blocks of various size into a goal that is at the edge of a square stage. At the beginning of each episode, the blocks, agents, and goals are randomly placed in the stage. When one of the blocks hits the goal, all the agents receive a group reward corresponding to the size of the block. Small blocks are +1, medium +2, and large +3. Large blocks require all three agents to push together to move at any reasonable speed, 2 blocks require two agents to collaborate, and small blocks can be pushed by a single agent. The episode ends when all blocks are pushed into the goal or when 1000 steps have been taken by the agents. A small time penalty of -0.0025 per timestep is given to the agents to encourage them to finish quickly. Agents' observe by casting 21 rays in a 180$^{\circ}$ arc front of them, similar to a LIDAR. An agent is given the distance to an object that the ray collides with, as well as if it is an agent, a wall, the goal, or the type of block. Two sets of rays are given, one that is high enough to see the walls and goal over the blocks and agents, and one that is at agent-level.


\textbf{Dungeon Escape.} This environment contains five agents, a dragon holding a key (green), two dragons that don't hold a key (pink), a portal, and a door in a square stage. At the beginning of each episode, the agents, the dragons, and the door and portal are randomly placed in the stage. The agents' receive a +1 group reward if at least one of them to exit the stage through the door. To do so, this agent must first have a key, which is dropped by the green dragon. In order to slay the dragon and make it drop the key, an agent needs to run into it, which will cause both the agent and the dragon to be removed from the environment. The green dragon slowly moves towards the portal, while the pink dragons will attempt to eat the closest agent. The episode ends if the green dragon reaches the portal or an agent with the key reaches the door. In order for the group of agents to receive a reward, at least one of the agents \emph{must learn to sacrifice itself} so that another can grab the key and go through the door, and one or more of them must learn to distract the pink dragons so that they do not attack the key-holding agent. Similarly to Cooperative Push Block, agents' observe by casting 15 rays in a 120$^{\circ}$ arc front of them. Unlike Cooperative Push Block, only one set of rays are used as there are less elements to obstruct them.


\textbf{Baton Pass.} In this environment, a single agent is spawned along side an orb and a button. The button can only be pressed if the orb has already been collected. Pressing the button will create a new orb and spawn a new agent. Only the most recently spawned agent can collect the orb or press the button. This forces each agent to first collect the orb, then press the button and finally stay out of the way of the newly spawned agents. Indeed the agents are large and can block each other. Every time an orb is collected the whole group gets a +1 reward, the game ends when 20 orbs have been collected. Each agent also has the possibility to be despawned by touching an exit zone. Doing so will not grant any rewards or penalties but will free available space for other agents. At each time step, the agents will receive a penalty of 0.000125 times the number of currently existing agents. This is to force the agents to finish the task faster and encourage agents to despawn once they can no longer collect orbs or press the button. The agents perceive the environment with one set of 13 raycasts plus information about their velocity and their capacity to press the button or collect orbs. The agents can move forward, backwards and rotate.


\textbf{Simple Spread.} This environment is taken directly from set of Multi-Agent Particle Environments created by~\cite{Lowe2017}. Agents are rewarded based on the minimum distance between each landmark (shown as dots) and any agent. Agents are penalized if they collide with other agents; the optimal strategy is to cover all the dots without colliding with each other. Agents observe the position of themselves, their teammates, and all 3 landmarks. All rewards are given as a group reward for the entire team of 3 agents. 

\section{Hyperparameters}
\label{sec:hyperparameters}
\subsection{Common Hyperparameters for Reinforcement Learning Tasks}

Table~\ref{tab:all_hyp} shows the hyperparameters used for all environments. As our implementations of PPO, MA-POCA and COMA all use the same policy and value clipping and entropy bonus mechanisms, all hyperparameters apply to all three algorithms. Network size parameters apply to both the critic and policy. In PPO, $\lambda$ is used for GAE whereas in MA-POCA and COMA it is used for $TD(\lambda)$.

\begin{table}[H]
\footnotesize
\begin{center}
\setlength{\tabcolsep}{1pt}
  \begin{tabular}{ c | c | c}
     & Dungeon Escape &  \\
     Hyperparameter & Baton Pass & Simple Spread  \\
      & Push Block &  \\\hline
    Minibatch Size & 1024 & 512 \\ 
    Buffer Size & 10240 & 5120 \\
    Epochs per Update & 3 & 3  \\
    Learning Rate & 0.0003 & 0.0003 \\
    Optimizer & Adam & Adam \\
    Entropy Bonus $\beta$ & 0.01 & 0.01 \\
    Clip Ratio $\epsilon$ & 0.2 & 0.2 \\ 
    $\lambda$ & 0.95& 0.95 \\
    Discount Factor $\gamma$ & 0.99 & 0.99 \\
    Hidden Units & 256 & 128 \\
    Fully Connected Layers & 2 & 2 \\
    Attention Entity Embedding Size\mbox{*} & 256 & 128 \\
    Attention Entity Embedding Layers\mbox{*} & 1 & 1 \\
    Number of Attention Heads\mbox{*} & 4 & 4  \\
  \end{tabular}
\end{center}
\caption{Hyperparameters for all algorithms for Collaborative Push Block, Baton Pass and Dungeon Escape (under the General Case column) and for Simple Spread. The hyperparameters marked with an \mbox{*} only apply to MA-POCA since PPO and COMA do not use an attention module.}
\label{tab:all_hyp}
\end{table}

\subsection{Hyperparameters for Computing the Mean}
For the fully connected network and for the self-attention networks. Hyperparameters that do not apply are denoted with a `` / ".
\begin{table}[H]
\begin{center}
  \begin{tabular}{ c | c | c}
    Hyperparameter & Fully Connected & Attention \\ \hline
    Minibatch Size &  500 &  500 \\ 
    Learning Rate & 0.001 & 0.001 \\
    Optimizer & Adam & Adam \\
    Hidden Units & 32 & / \\
    Layers & 2 & / \\
    Entity Embedding Size & / & 32 \\
    Entity Embedding Layers & / & 1 \\
    Number of Attention Heads & / & 4 \\
  \end{tabular}
\end{center}
\caption{Hyperparameters for fully connected network and for self-attention network for the Compute the Mean task}
\label{tab:fc_hyp}
\end{table}

\end{document}